\documentclass[conference]{IEEEtran}
\IEEEoverridecommandlockouts
\usepackage[T1]{fontenc}
\usepackage[utf8]{inputenc}
\usepackage{amsmath,amssymb}
\usepackage{graphicx}
\usepackage{booktabs}
\usepackage{cite}
\usepackage{xcolor}
\usepackage[hidelinks]{hyperref}
\usepackage{url}
\newcommand{\orcid}[1]{\href{https://orcid.org/#1}{ORCID: #1}}

\begin{document}

\title{Visibility Graph Representations of Word Embeddings for Text Classification\thanks{Preprint. This paper is a conference-format version of a chapter submitted to the edited volume \emph{Advances in Artificial Intelligence and Data Science: Methods, Systems, and Applications} (Karab\"uk University Press). It extends the authors' 2022 congress paper and the first author's M.S. thesis (see Acknowledgment).}}

\author{\IEEEauthorblockN{Elif Doruk\-ba\c{s}{\i}}
\IEEEauthorblockA{\textit{Department of Computer Engineering}\\
\textit{Faculty of Computer and Information Sciences}\\
\textit{Karab\"uk University}, Karab\"uk, T\"urkiye\\
elifkabullar@karabuk.edu.tr\\
\orcid{0000-0003-3206-5784}}
\and
\IEEEauthorblockN{\.Ilker T\"urker\textsuperscript{*}}
\IEEEauthorblockA{\textit{Department of Computer Engineering}\\
\textit{Faculty of Computer and Information Sciences}\\
\textit{Karab\"uk University}, Karab\"uk, T\"urkiye\\
iturker@karabuk.edu.tr\\
\orcid{0000-0001-7577-4658}}
}

\maketitle

\begingroup\renewcommand\thefootnote{}\footnotetext{\textsuperscript{*}Corresponding author: \.Ilker T\"urker (iturker@karabuk.edu.tr).}\endgroup

\begin{abstract}
Text classifiers depend on representations that expose the distinctions between classes in a form the learning model can use, and a restricted vocabulary changes which representation is most effective. This paper studies natural visibility graphs as an intermediate representation for news topic classification under controlled vocabulary budgets. Word embeddings are averaged into a fixed-length document vector, the vector is read as an ordered numerical series, and its natural visibility graph is passed to a classifier either as an ordered node-degree vector or as an adjacency matrix. On the four-class AG News benchmark, chi-square selection retains 100 to 1,500 words, and the graph inputs are compared with direct embedding-based representations from continuous bag-of-words, skip-gram, GloVe, FastText and BERT. With a fully connected classifier, the degree representation is the most accurate configuration from 700 selected words onward; it reaches its highest accuracy of 91.24\% at 1,000 words and exceeds the BERT-based baseline by 0.29 percentage points at 1,500 words. In a separate convolutional comparison, the adjacency input reaches 86.16\% at 1,000 words, against 83.12\% for the strongest static embedding. The results indicate that visibility-derived features can be competitive once the selected vocabulary provides adequate coverage. We discuss the roles of graph compression, coordinate ordering and classifier choice, and the controls needed to establish the reliability of the small differences observed.
\end{abstract}

\begin{IEEEkeywords}
text classification, visibility graph, word embeddings, chi-square feature selection, node degree, news categorization
\end{IEEEkeywords}

\section{Introduction}
Automatic text classification assigns documents to predefined classes and supports news organisation, information filtering and sentiment analysis~\cite{gasparetto2022,khurana2023}. The classifier alone does not define the learning problem: the representation determines which properties of a document are available to the decision function. Vocabulary-based representations expose term usage, word embeddings supply distributed semantic information, and contextual encoders make the representation of a token depend on its context. Reviews of the field discuss these families together with the assumptions each of them makes~\cite{gasparetto2022,khurana2023}.

News categorisation is a natural test bed for these choices. A short news item may mention a company, a scientific development, an international event or a public policy, and the annotated topic depends on which of these dominates. Individual terms are therefore informative but not decisive, and a representation must aggregate their evidence so that the main content of the document is emphasised. The AG News benchmark provides four balanced topic categories and a standard training and test partition for studying this task~\cite{zhang2015}.

Vocabulary size is a practical design constraint. A small vocabulary yields a compact lexical feature space but may discard terms that separate closely related topics, whereas a large vocabulary improves coverage at the cost of redundant or weakly informative evidence. The vocabulary budget should therefore be studied jointly with the representation. Throughout this paper, the selected vocabulary size $K$ denotes the number of retained word types; it is distinct from document length, from the embedding dimension and from the number of graph vertices.

Distributed representations address some shortcomings of literal word matching, but a stronger encoder does not by itself settle the representation question, and transformer-based classification raises concerns of cost, input handling and reliability~\cite{fields2024}. Graphs offer a complementary view. Text graphs usually encode word co-occurrence, linguistic dependencies or document--term relations, and their usefulness depends on how nodes and edges are defined~\cite{wang2024}. The present approach instead applies a visibility rule to the ordered numerical representation of each document and uses the resulting connectivity either as a node-degree vector or as the input of an adjacency-based classifier. This is a numerical rather than a linguistic construction: a visibility edge records an unobstructed geometric relation between scalar values at ordered positions, not a semantic relation between words, and a network trained on graph degrees is not a graph neural network merely because its input is derived from a graph.

The study addresses three questions. How do direct embedding-based representations behave as the vocabulary budget grows? Under which budgets do visibility-derived features become competitive? How does the ranking change when a fully connected classifier is replaced by a convolutional one? The contributions are as follows:
\begin{itemize}
\item A per-document representation that converts an embedding-based document vector into a natural visibility graph and supplies it to standard neural classifiers as an ordered degree vector or as an adjacency matrix.
\item A controlled comparison on AG News of five embedding-based baselines and the graph inputs across fifteen chi-square vocabulary budgets and two classifier families.
\item An analysis of the conditions under which the graph inputs are competitive, together with the structural properties, limitations and validation controls that bear on interpreting small differences.
\end{itemize}
Section~\ref{sec:related} reviews related work, Section~\ref{sec:method} describes the representation framework, Section~\ref{sec:setup} presents the experimental setting, Section~\ref{sec:results} reports and discusses the results, and Section~\ref{sec:conclusion} concludes the paper.

\section{Related Work}\label{sec:related}
\subsection{From Lexical Features to Distributed Representations}
Single-label classification maps a document $d$ to one label $y$ from a finite set. A representation function maps $d$ to a numerical input, and a classifier learns the mapping from that input to the label; the two stages can be separated to study the representation, provided that the task, data partition and evaluation criteria are held fixed. Frequency-based representations assign lexical identities to their coordinates but are sparse in high-dimensional vocabularies, which motivates supervised feature selection that ranks terms by their association with the classes~\cite{yang1997}.

Word2Vec learns dense word vectors with the continuous bag-of-words (CBOW) and skip-gram objectives~\cite{mikolov2013}; GloVe learns them from global co-occurrence statistics~\cite{pennington2014}; and FastText composes word vectors from character n-grams so that words with similar spelling share information~\cite{bojanowski2017}. BERT pre-trains a bidirectional transformer to produce context-dependent token representations~\cite{devlin2019}. For document classification, word vectors must be aggregated or processed as a sequence, so the embedding family and the document-construction rule are distinct components of a method. Preprocessing alone measurably changes embedding quality~\cite{rahimi2023}, and comparisons of static and contextual embeddings with neural classifiers report strong representation--classifier interactions~\cite{sabbeh2023}.

\subsection{Graphs for Text Classification}
Graph-based classifiers make relations explicit; nodes may be words, tokens, sentences or documents, and edges may encode co-occurrence, syntax, semantic similarity or corpus associations. TextGCN builds a single corpus graph of word co-occurrence and document--word relations and learns with graph convolution~\cite{yao2019}, which makes it transductive. Onan fuses contextual node embeddings and BERT within a hierarchical graph~\cite{onan2023}, and Donabauer and Kruschwitz build token-level graphs from pre-trained language model tokenisation and embeddings~\cite{donabauer2025}. These studies show that relational and contextual information can be combined, and also why the effects of graph construction, embedding quality and learning architecture must be separated when a pipeline is evaluated.

\subsection{Visibility Graphs}
Lacasa et al.\ map a time series to a complex network by connecting pairs of observations that can see each other over all intermediate values~\cite{lacasa2008}. Wen et al.\ review the natural, horizontal and further visibility constructions and their use in prediction and classification~\cite{wen2022}. Visibility features have been used for sound through the connectogram representation~\cite{turker2022} and for electrocardiogram classification with weighted visibility graphs~\cite{kutluana2024}; in these domains the series axis has a physical meaning that does not carry over to embedding coordinates. Ghadiri et al.\ combine visibility-graph features of voice signals with transcription-based text features for depression detection~\cite{ghadiri2023}, which is complementary to, but methodologically different from, the document-vector transformation studied here. Table~\ref{tab:related} summarises this positioning: the present study does not propose a message-passing architecture, but asks whether visibility connectivity and its degree summary provide useful inputs to conventional neural classifiers under limited vocabularies.

\begin{table}[t]
\caption{Related approaches and the position of the present study.}
\label{tab:related}
\centering\footnotesize
\begin{tabular}{@{}p{2.1cm}p{2.7cm}p{2.6cm}@{}}
\toprule
Approach & Graph or representation & Learning setting\\
\midrule
TextGCN~\cite{yao2019} & Corpus word--document relations & Graph convolution\\
Onan~\cite{onan2023} & Hierarchical linguistic graph & Contextual fusion and graph learning\\
Donabauer and Kruschwitz~\cite{donabauer2025} & Contextual token graph & Graph-based text classification\\
Ghadiri et al.~\cite{ghadiri2023} & Voice visibility graph plus text & Multimodal fusion\\
\textbf{This study} & \textbf{Visibility graph of a numerical document series} & \textbf{Degree or adjacency inputs}\\
\bottomrule
\end{tabular}
\end{table}

\section{Method}\label{sec:method}
\subsection{Framework}
\begin{figure}[t]
\centering
\includegraphics[width=\columnwidth]{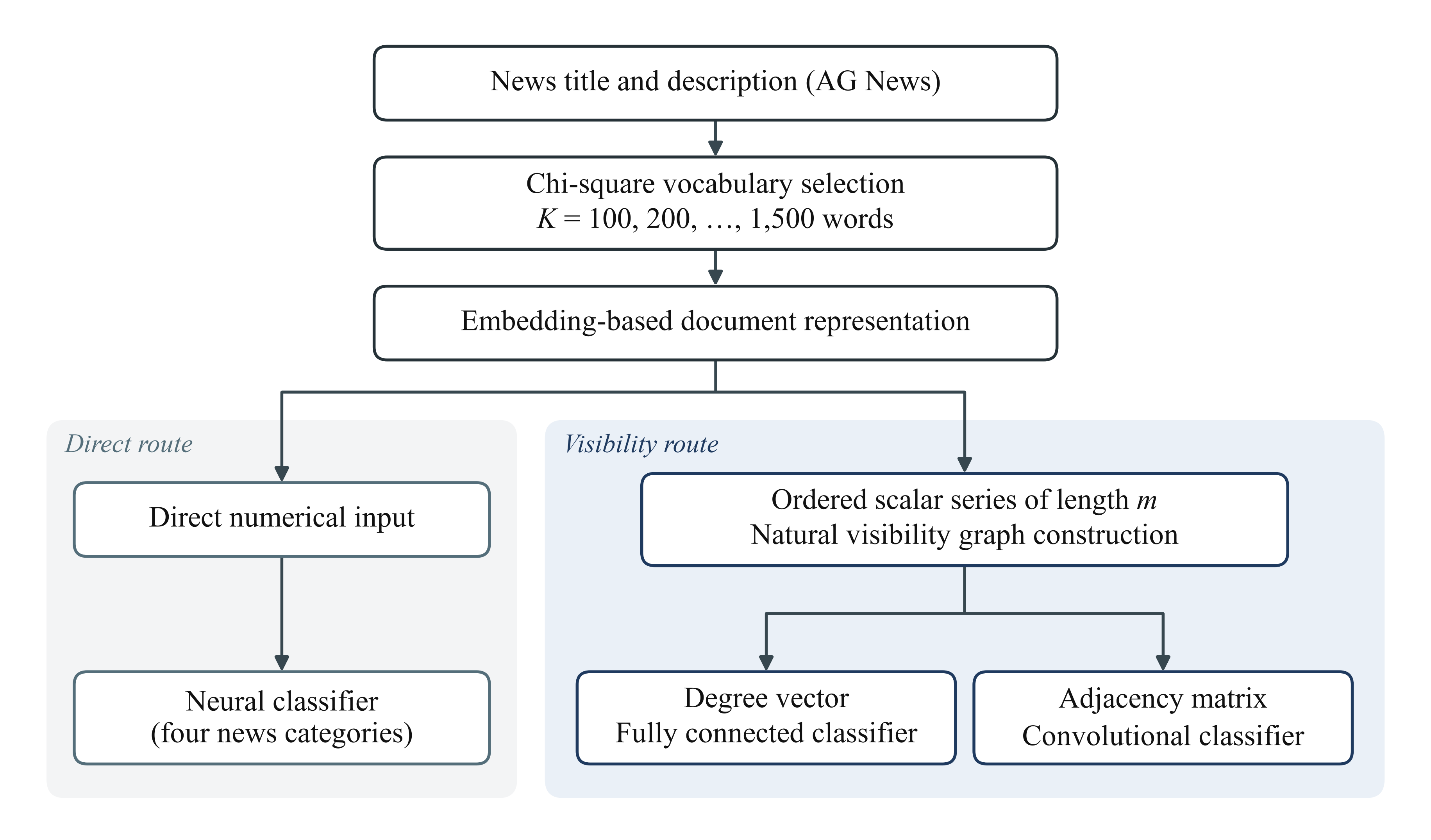}
\caption{Classification with direct and visibility-derived numerical representations. The vocabulary size $K$ and the length $m$ of the numerical series are distinct quantities.}
\label{fig:framework}
\end{figure}

Fig.~\ref{fig:framework} shows the processing framework. Each news title is concatenated with its description, a chi-square criterion selects a vocabulary of $K$ words, and an embedding-based document representation is formed. In the direct route this representation is classified as it is. In the visibility route it is read as an ordered scalar series of length $m$ and converted into a natural visibility graph, whose connectivity is retained as an adjacency matrix or compressed into an ordered node-degree vector. The transformation makes relative structural properties of the numerical representation explicit before classification; it adds no new observations or labels.

\subsection{Document Preparation and Vocabulary Selection}
AG News items contain a category identifier, a title and a description; the title and description are concatenated into a single text field, and the classifier predicts one label for the combined item. The study considers $K = 100, 200, \dots, 1{,}500$ selected word types. Selection uses the chi-square statistic between term occurrence and class membership, whose contingency-table form is
\begin{equation}
\chi^2(w) = \sum_{r}\sum_{c} \frac{\left(O_{rc} - E_{rc}\right)^2}{E_{rc}},
\label{eq:chi2}
\end{equation}
where $w$ is a candidate word, $r$ indexes its occurrence state, $c$ indexes the class, $O_{rc}$ is the observed count and $E_{rc}$ the count expected under independence. Larger values indicate a stronger departure from independence; the statistic measures association, not meaning. Because the criterion uses class labels, it belongs to the training procedure and must be fitted on training data only.

\subsection{Numerical Series and Visibility Graph Construction}
Each document is represented by the component-wise mean of the word-embedding vectors available for its words within the selected vocabulary; a document without any such vector is mapped to the zero vector. With the 100-dimensional embeddings used in the graph experiments, every document therefore yields an ordered series of $m = 100$ values whose positions are embedding coordinates, independently of $K$.

\begin{figure}[t]
\centering
\includegraphics[width=\columnwidth]{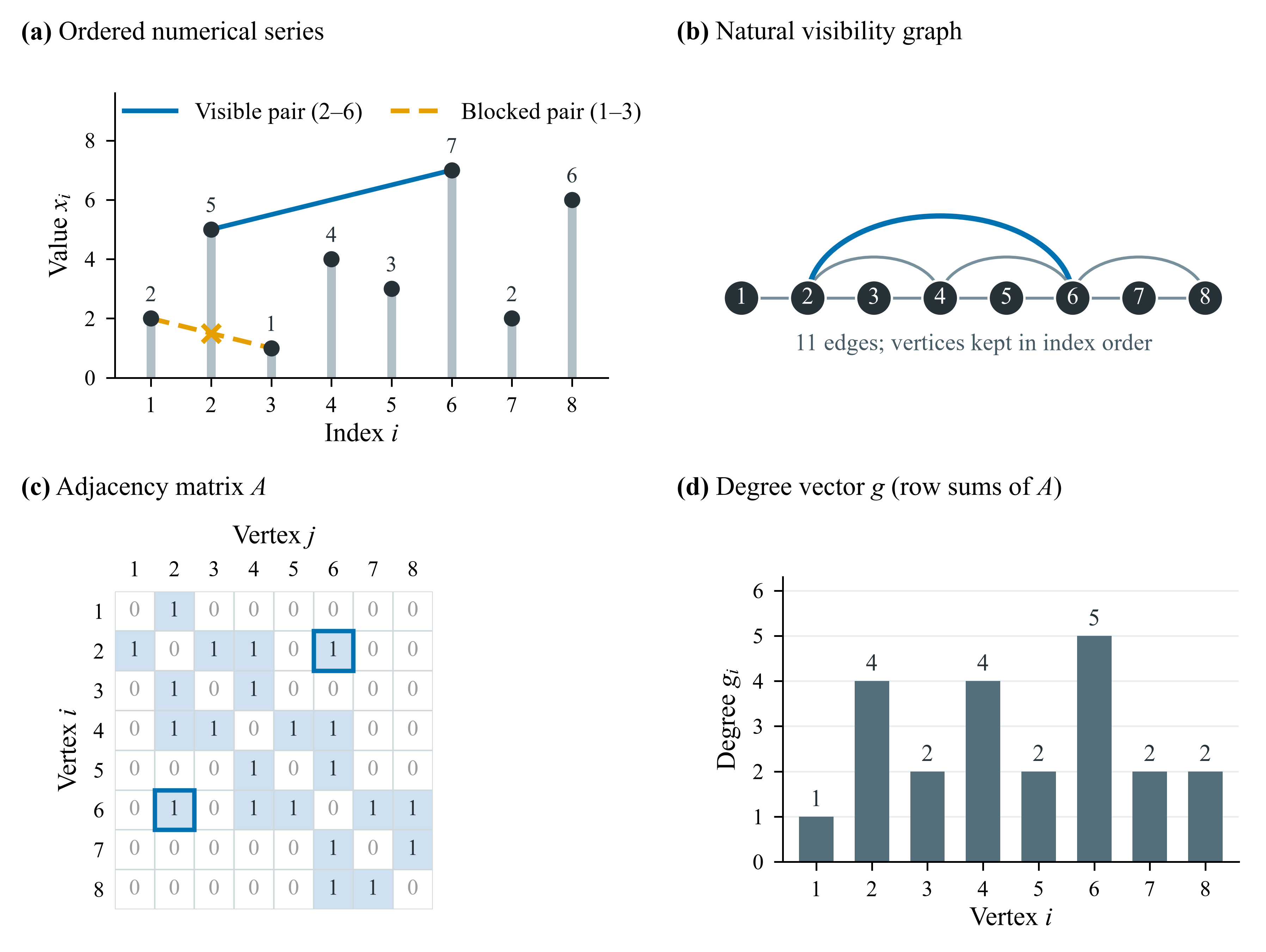}
\caption{Natural visibility construction for an eight-value series: (a) the numerical input with one visible and one blocked pair; (b) all eleven graph edges; (c) the symmetric adjacency matrix; (d) node degrees obtained as row sums.}
\label{fig:nvg}
\end{figure}

Let $\mathbf{x} = (x_1, \dots, x_m)$ be the ordered series. Each position is a vertex, and vertices $i$ and $j$ with $i < j$ are connected under natural visibility if every intermediate value lies strictly below the segment that joins them:
\begin{equation}
x_k < x_i + \left(x_j - x_i\right)\frac{k - i}{j - i}, \qquad i < k < j.
\label{eq:nvg}
\end{equation}
Adjacent positions are always connected. The resulting undirected, unweighted adjacency matrix $A$ has $A_{ij} = 1$ for visible pairs, $A_{ij} = 0$ otherwise and $A_{ii} = 0$. Fig.~\ref{fig:nvg} illustrates the construction. The graphs were generated with the ts2vg library~\cite{ts2vg} and processed with NetworkX.

Visibility is a deliberate, lossy structural transformation. Adding a constant to all values or rescaling them by a positive factor leaves every inequality in~\eqref{eq:nvg} unchanged, so different series can produce the same graph. A permutation of coordinates, in contrast, can change the graph even though the set of values is unchanged. The representation is therefore neither invariant to ordering nor reversible. For embeddings, neighbouring coordinates are not neighbouring words and carry no linguistic adjacency, so the usefulness of the edges must be established by predictive results and ordering controls rather than by a semantic interpretation.

\subsection{Graph Read-outs and Classifiers}
The degree read-out summarises connectivity as
\begin{equation}
g_i = \sum_{j=1}^{m} A_{ij}, \qquad i = 1, \dots, m,
\label{eq:degree}
\end{equation}
and the vector $\mathbf{g} = (g_1, \dots, g_m)$ is normalised to $[0, 1]$ and passed to a fully connected artificial neural network. Degrees keep the relative connectivity of each position but not the identity of its neighbours, so different graphs can share a degree vector. The adjacency read-out keeps the full $100 \times 100$ matrix, i.e.\ $10{,}000$ entries against $100$ for the degree vector, and passes it to a convolutional classifier. Conventional convolution operates on the grid order of the matrix, whereas graph convolution propagates information along graph neighbourhoods~\cite{kim2014}; the convolutional classifier used here is therefore not a message-passing graph model. Following the original implementation~\cite{thesis}, the fully connected classifier is a multilayer perceptron with rectified linear hidden layers, dropout and a four-way softmax output, and the convolutional classifier applies two-dimensional convolution and max pooling to the adjacency matrix before dense layers and a softmax output. The direct baselines use CBOW, skip-gram, GloVe, FastText and BERT representations.

\section{Experimental Setting}\label{sec:setup}
\subsection{Benchmark and Comparison Design}
The experiments use the four-class AG News benchmark~\cite{zhang2015}, with 30,000 training and 1,900 test documents in each of the World, Sports, Business and Science \& Technology categories (Table~\ref{tab:data}). The fully connected comparison covers six representations, five embedding-based inputs and the graph-degree input, over fifteen budgets from 100 to 1,500 words. A separate convolutional comparison covers the graph adjacency input, skip-gram, CBOW and GloVe over ten budgets from 100 to 1,000 words. All scores come from the original experiments~\cite{thesis,kabullar2022}; no scores are combined from unrelated studies.

\begin{table}[t]
\caption{AG News benchmark distribution.}
\label{tab:data}
\centering\footnotesize
\begin{tabular}{@{}lrr@{}}
\toprule
Category & Training & Test\\
\midrule
World & 30{,}000 & 1{,}900\\
Sports & 30{,}000 & 1{,}900\\
Business & 30{,}000 & 1{,}900\\
Science and Technology & 30{,}000 & 1{,}900\\
\midrule
Total & 120{,}000 & 7{,}600\\
\bottomrule
\end{tabular}
\end{table}

The available records do not fully document the validation split, repeated runs or the training-only fitting of vocabulary selection. The reported scores are therefore treated as point estimates of single experiments; they do not quantify variability across random seeds, and small differences are interpreted accordingly.

\subsection{Evaluation Measure}
Accuracy is the primary measure. For $N$ evaluated documents,
\begin{equation}
\mathrm{Accuracy} = \frac{1}{N}\sum_{n=1}^{N} \mathbb{I}\left[\hat{y}_n = y_n\right],
\label{eq:acc}
\end{equation}
where $y_n$ and $\hat{y}_n$ are the true and predicted categories and $\mathbb{I}[\cdot]$ is the indicator function. Because the classes are balanced, accuracy is a useful first summary, but it does not show whether errors concentrate in particular classes. The averaging convention of the original F-scores was not recorded, so they are not reinterpreted here. Differences between accuracies are reported in percentage points (pp).

\subsection{Implementation}
The experiments were implemented in Python with Jupyter Notebook, using scikit-learn and NLTK for text processing, NumPy and pandas for data handling, gensim for embeddings, TensorFlow for the neural classifiers, ts2vg and NetworkX for graph construction and seaborn for visualisation. This paper reorganises and analyses the recorded experiments and contains no new classification runs.

\section{Results and Discussion}\label{sec:results}
\subsection{Fully Connected Classification Across Vocabulary Budgets}
\begin{figure*}[t]
\centering
\includegraphics[width=0.78\textwidth]{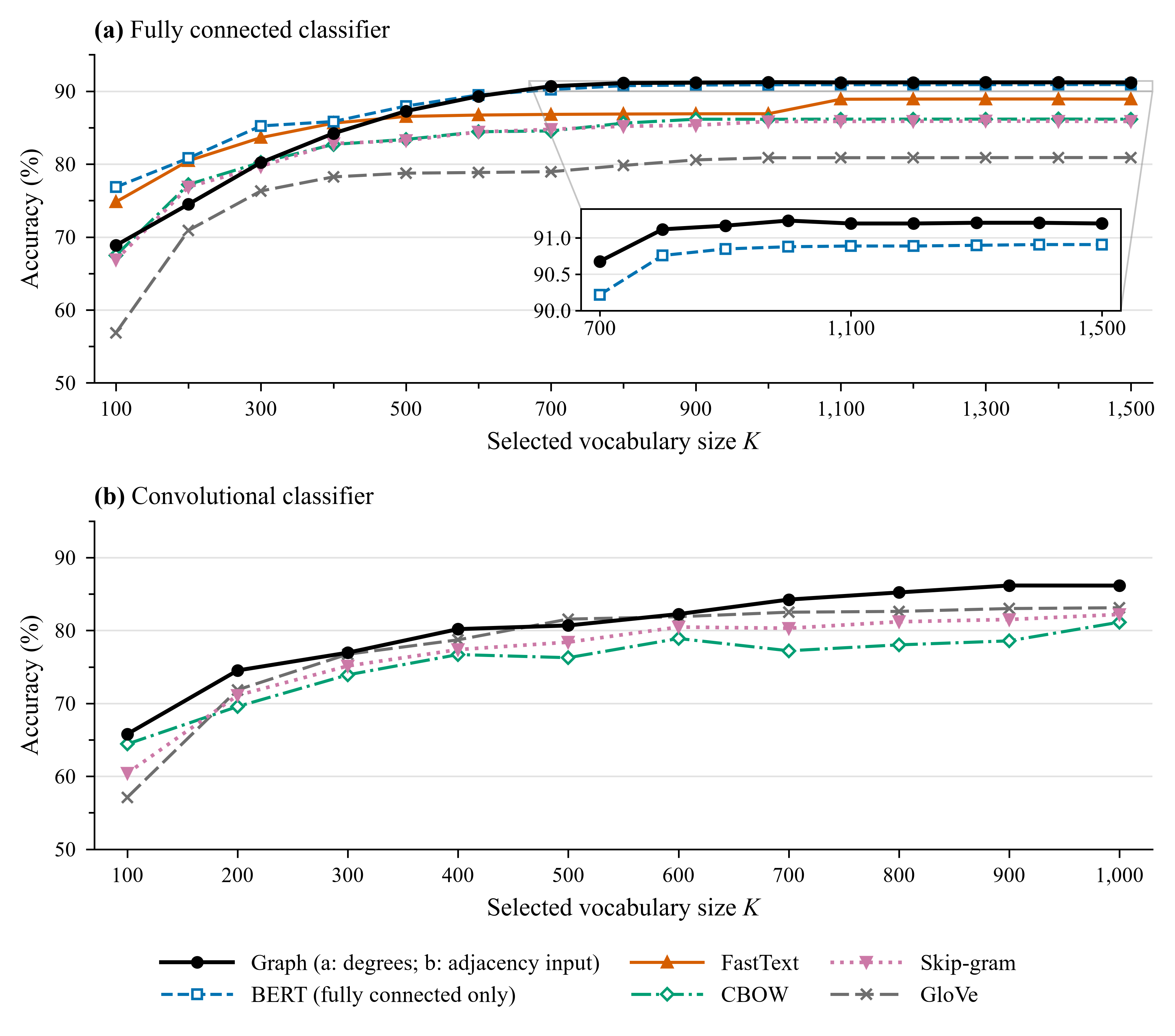}
\caption{Accuracy across selected vocabulary budgets. (a) Fully connected classifier with direct embeddings and graph degrees; the inset magnifies $K = 700$--$1{,}500$ for graph degrees and BERT. (b) Convolutional classifier with static embeddings and the graph adjacency input.}
\label{fig:results}
\end{figure*}

Table~\ref{tab:fc} lists the full accuracy sweep for the direct embeddings and the graph degrees, and Fig.~\ref{fig:results}a plots the same values. At 100 selected words BERT reaches 76.85\%, FastText 74.82\% and the graph degrees 68.86\%, so the graph transformation has no advantage with the smallest vocabulary. The degree accuracy rises steeply to 87.27\% at 500 words, where BERT still leads with 87.96\%, and at 600 words the two differ by only 0.17~pp (89.31\% against 89.48\%).

\begin{table}[t]
\caption{Fully connected accuracy (\%) across vocabulary budgets $K$. The best value in each row is in bold. Values from~\cite{thesis,kabullar2022}.}
\label{tab:fc}
\centering\footnotesize
\setlength{\tabcolsep}{3.2pt}
\begin{tabular}{@{}rrrrrrr@{}}
\toprule
$K$ & GloVe & FastText & Skip-gram & CBOW & BERT & Degrees\\
\midrule
100 & 56.88 & 74.82 & 66.88 & 67.50 & \textbf{76.85} & 68.86 \\
200 & 70.92 & 80.46 & 76.87 & 77.21 & \textbf{80.82} & 74.50 \\
300 & 76.32 & 83.64 & 79.70 & 80.24 & \textbf{85.24} & 80.22 \\
400 & 78.25 & 85.62 & 82.82 & 82.70 & \textbf{85.84} & 84.22 \\
500 & 78.76 & 86.54 & 83.23 & 83.39 & \textbf{87.96} & 87.27 \\
600 & 78.85 & 86.73 & 84.41 & 84.44 & \textbf{89.48} & 89.31 \\
700 & 78.96 & 86.82 & 84.77 & 84.55 & 90.22 & \textbf{90.68} \\
800 & 79.82 & 86.87 & 85.20 & 85.62 & 90.76 & \textbf{91.12} \\
900 & 80.58 & 86.90 & 85.33 & 86.16 & 90.85 & \textbf{91.17} \\
1{,}000 & 80.87 & 86.91 & 85.85 & 86.15 & 90.88 & \textbf{91.24} \\
1{,}100 & 80.88 & 88.90 & 85.86 & 86.17 & 90.89 & \textbf{91.20} \\
1{,}200 & 80.88 & 88.92 & 85.88 & 86.17 & 90.89 & \textbf{91.20} \\
1{,}300 & 80.89 & 88.93 & 85.90 & 86.17 & 90.90 & \textbf{91.21} \\
1{,}400 & 80.90 & 88.93 & 85.88 & 86.18 & 90.91 & \textbf{91.21} \\
1{,}500 & 80.90 & 88.91 & 85.86 & 86.16 & 90.91 & \textbf{91.20} \\
\bottomrule
\end{tabular}
\end{table}

The graph degrees first exceed BERT at 700 words (90.68\% against 90.22\%) and remain the most accurate configuration at every larger budget. They reach their highest accuracy, 91.24\%, at 1,000 words, where BERT obtains 90.88\%. At 1,500 words the degrees obtain 91.20\%, BERT 90.91\%, FastText 88.91\%, CBOW 86.16\%, skip-gram 85.86\% and GloVe 80.90\%. The 0.29~pp margin over BERT at 1,500 words is a descriptive difference between single runs of the recorded configurations. It does not establish statistical significance, superiority over every BERT fine-tuning strategy or an advantage on other corpora, and it is small compared with the changes produced by the initial expansion of the vocabulary.

\subsection{Vocabulary Coverage and Saturation}
From 100 to 500 selected words the degree accuracy increases by 18.41~pp, against 11.11~pp for BERT. The graph representation thus benefits most once the selected vocabulary is large enough to separate the topics; the gain may stem from lexical coverage, from the stability of the averaged document vector, from the structural features themselves or from their interaction with the classifier, and it does not imply that the transformation creates semantic information. Beyond 1,000 words the sweep is nearly flat: the degrees vary between 91.20\% and 91.24\% and BERT between 90.88\% and 90.91\%, so the largest tested budget is not necessarily the best one. FastText is competitive with small budgets but remains below BERT, by 2.03~pp at 100 words and 1.42~pp at 500 words; its accuracy jumps from 86.91\% at 1,000 words to 88.90\% at 1,100 words, a discontinuity that should be checked against the training records before it is interpreted. A plateau identifies a candidate range for validation, not a final model choice; choosing the budget by repeatedly inspecting test accuracy would bias the reported endpoint.

\subsection{Convolutional Classification}
Table~\ref{tab:cnn} and Fig.~\ref{fig:results}b report the convolutional comparison. The graph adjacency input is the most accurate configuration at nine of the ten budgets; GloVe leads only at 500 words (81.56\% against 80.69\%), and at 300 words the graph input is ahead by only 0.21~pp (76.96\% against 76.75\%). At 1,000 words the graph input reaches 86.16\%, 3.04~pp above GloVe (83.12\%), with skip-gram at 82.20\% and CBOW at 81.14\%; its highest value, 86.17\%, occurs at 900 words.

\begin{table}[t]
\caption{Convolutional accuracy (\%) across vocabulary budgets $K$. ``Graph'' denotes the $100\times100$ adjacency matrix of the visibility graph. Best value per row in bold. Values from~\cite{thesis}.}
\label{tab:cnn}
\centering\footnotesize
\begin{tabular}{@{}rrrrr@{}}
\toprule
$K$ & GloVe & Skip-gram & CBOW & Graph\\
\midrule
100 & 57.12 & 60.43 & 64.43 & \textbf{65.80} \\
200 & 71.86 & 71.10 & 69.57 & \textbf{74.52} \\
300 & 76.75 & 75.14 & 73.92 & \textbf{76.96} \\
400 & 78.72 & 77.38 & 76.70 & \textbf{80.19} \\
500 & \textbf{81.56} & 78.40 & 76.27 & 80.69 \\
600 & 81.89 & 80.47 & 78.90 & \textbf{82.25} \\
700 & 82.51 & 80.31 & 77.22 & \textbf{84.24} \\
800 & 82.62 & 81.20 & 78.03 & \textbf{85.22} \\
900 & 83.01 & 81.51 & 78.58 & \textbf{86.17} \\
1{,}000 & 83.12 & 82.20 & 81.14 & \textbf{86.16} \\
\bottomrule
\end{tabular}
\end{table}

The convolutional ranking differs from the fully connected one. GloVe is the weakest direct embedding with the fully connected classifier but is ahead of both Word2Vec variants in the convolutional comparison at 1,000 words, which argues against treating the performance of an embedding family as independent of input construction and classifier. The fully connected degree input is more accurate than the convolutional adjacency input at every common budget except 200 words (74.50\% against 74.52\%), by up to 7.06~pp at 600 words. Because the read-out and the architecture change together, this comparison does not isolate the effect of degree compression; a matched experiment would vary one factor at a time.

\subsection{Structural Interpretation and Practical Implications}
The results support numerical connectivity as a potentially useful intermediate feature. The degree vector records how much of the series each ordered position can see, which changes the input geometry while retaining a relative pattern after the partial removal of amplitude information. Possible explanations include a regularising effect, an emphasis on class-specific structure and a better match with the classifier; these are hypotheses to be tested against controls, not established mechanisms. Compactness and efficiency must also be distinguished: the degree vector has 100 entries against 10,000 for the dense adjacency matrix, but this reduction concerns only the final input, and end-to-end cost also depends on text preparation, embedding, graph construction, batching and model size, which require direct measurement on the target hardware. For deployment, vocabulary coverage, the handling of out-of-vocabulary words, the ordering of numerical inputs and the stability of the graph under a change of embedding model are additional design factors.

\subsection{Limitations}
The evaluation uses a single balanced English benchmark and does not cover imbalanced categories, other languages, sentiment tasks, long documents or shifting news distributions. Each configuration was run once, so repeated runs with different seeds and paired prediction analyses are required before small differences can be considered reliable, and accuracy alone hides class-level precision, recall and confusion patterns. The conclusions also depend on the mapping from embeddings to the ordered series, the graph variant, the normalisation, the pre-trained checkpoints and the training settings, all of which should be documented in a released implementation. The original study also contains a diagonal-only feature configuration; since a loop-free adjacency matrix has a zero main diagonal, that configuration requires a precise definition before its scores can be interpreted, and it is therefore not reported here. Useful controls include fixing the document vectors and the classifier while varying only the visibility features against a size-matched structural control, fixing the graph input while varying the classifier, and permuting the coordinate order.

\section{Conclusion}\label{sec:conclusion}
This paper examined natural visibility graphs as an intermediate numerical representation for text classification, linking distributed word features and supervised vocabulary selection to geometric graph construction, degree compression and neural learning. In the recorded AG News experiments, graph degrees reached 91.20\% accuracy with 1,500 selected words against 90.91\% for a BERT-based baseline, with a maximum of 91.24\% at 1,000 words; in the convolutional comparison the adjacency input reached 86.16\% at 1,000 words against 83.12\% for GloVe. Visibility construction provides an explicit connectivity rule and degree extraction a compact summary of it, neither of which implies a semantic interpretation of the edges or a general advantage over direct embeddings. The nearly flat degree accuracy between 1,000 and 1,500 words gives practitioners a candidate range for validation. Future work will provide a complete public implementation with training-only feature selection, repeated runs and paired tests, matched ablations and coordinate-order controls, and evaluations on other languages and document types with class-level error analysis.

\section*{Acknowledgment}
This work extends the authors' congress paper presented at the IV International Istanbul Modern Scientific Research Congress (Istanbul, T\"urkiye, 16--18 December 2022)~\cite{kabullar2022} and the first author's M.S. thesis completed at Karab\"uk University in 2023~\cite{thesis}. Claude (Opus 5.5) was used to assist with organising the manuscript, preparing figures and tables from existing results and adapting the text to this format; it was not used to run the reported experiments. The authors take full responsibility for the content.

\section*{Data Availability}
AG News is publicly distributed with its original publication~\cite{zhang2015} (\url{https://github.com/zhangxiangxiao/Crepe}). Experimental code and document-level predictions are not publicly archived at the time of writing.

\end{document}